\documentclass[letterpaper]{article} 
\usepackage{aaai20}  
\usepackage{times}  
\usepackage{helvet} 
\usepackage{courier}  
\usepackage[hyphens]{url}  
\usepackage{graphicx} 
\usepackage{graphicx}  
\nocopyright
\usepackage{epsfig}
\usepackage{amsmath}
\usepackage{amssymb}
\usepackage{url}
\usepackage{xcolor}

\title{Temporally Grounding Language Queries in Videos by \\Contextual Boundary-aware Prediction}
\author{Jingwen Wang\qquad Lin Ma\qquad Wenhao Jiang\\ \\
Tencent AI Lab
\\
{\tt\small$\lbrace$jaywongjaywong, forest.linma, cswhjiang$\rbrace$@gmail.com}
}

\begin{document}

\maketitle

\begin{abstract}
   The task of temporally grounding language queries in videos is to temporally localize the best matched video segment corresponding to a given language (sentence). It requires certain models to simultaneously perform visual and linguistic understandings. Previous work predominantly ignores the precision of segment localization. Sliding window based methods use predefined search window sizes, which suffer from redundant computation, while existing anchor-based approaches fail to yield precise localization. We address this issue by proposing an end-to-end boundary-aware model, which uses a lightweight branch to predict semantic boundaries corresponding to the given linguistic information. To better detect semantic boundaries, we propose to aggregate contextual information by explicitly modeling the relationship between the current element and its neighbors. The most confident segments are subsequently selected based on both anchor and boundary predictions at the testing stage. The proposed model, dubbed \underline{C}ontextual \underline{B}oundary-aware \underline{P}rediction (CBP), outperforms its competitors with a clear margin on three public datasets. All codes are available on \texttt{\textcolor{blue}{\url{https://github.com/JaywongWang/CBP}}}.
\end{abstract}

\setcounter{secnumdepth}{1} 
\section{Introduction}
\label{introduction}

Videos are increasingly popular in the social network. As most videos contain both activities of interest and complicated background content, temporal activity localization is of key importance for video analysis. Recently, the task of temporally grounding language queries in videos has been attracting research interest from the vision community \cite{gao2017tall,hendricks2017localizing}. The task aims to localize the activity of interest corresponding to a language query. This task is challenging because both videos and sentences need to be deeply incorporated to differentiate fine-grained details of different video segments and to perform segment localization. In this paper, we identify and tackle the main challenge on this task, namely, how to improve the localization precision of the desired segment given a language query.

\begin{figure}[t]
\begin{center}
\includegraphics [width=1.0\linewidth]{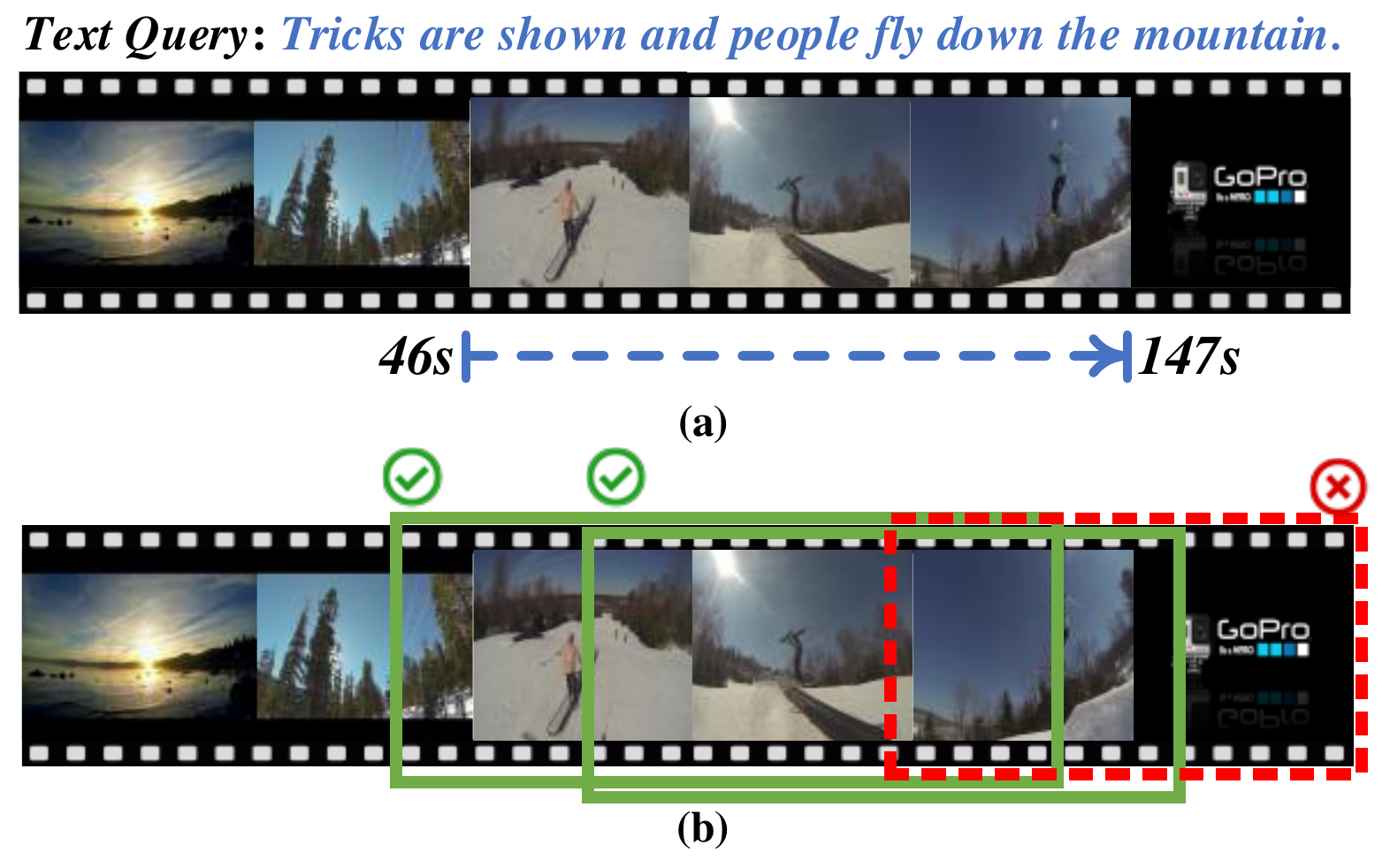}
\vspace{-15pt}
\caption{(a) The task of temporally grounding language queries in videos. (b) Positive and negative training segments defined in anchor-based approaches given the sentence query in (a).}
\label{figure_introduction}
\end{center}
\vspace{-5pt}
\end{figure}

Prior work predominantly ignores the precision of segment boundaries. Sliding window based methods scan the video by predefined windows of different sizes \cite{gao2017tall,hendricks2017localizing,liu2018cross,wu2018multi,liu2018attentive,ge2019mac,xu2019multilevel}. Because the desired segments are of varied durations, these methods cannot guarantee the complete coverage of all segments, and thus tend to produce inaccurate temporal boundaries. Other research tried to avoid this problem by designing single-stream models \cite{buch2017sst,chen2018temporally} using LSTMs. Although LSTMs effectively aggregate video information, the thresholding of positive and negative samples loses boundary information. As shown in Figure \ref{figure_introduction} (b), segments overlapped with the ground truth more than a predefined threshold (e.g., 0.5) are all labeled as positive samples during training stage. Therefore, the model could be confused to localize the best matched segment at prediction. A complementary approach to improve the precision of localization is to add a location offset regression branch to the anchor-based approaches \cite{gao2017tall,xu2019multilevel,ge2019mac,liu2018attentive}. However, the added offset regression could fail when the model is unable to localize the best anchor, since the calculated offsets need to be added to the predicted anchor to generate final grounding time stamp (See Table \ref{table_ablation_performance} for comparison). 

To improve temporal grounding precision, we propose a novel model that jointly predicts temporal anchors and boundaries at each time step, with a small computation overhead. At prediction stage, the anchors are modulated by boundary scores to generate boundary-aware grounding results. To detect semantic boundaries more accurately, contextual information is adaptively integrated into our architecture. As shown in Figure \ref{figure_introduction}, the activity ``fly down the mountain'' exhibits different visual appearance compared to the background content. The activity is better localized with the aid of its surrounding information. To this end, we propose a self attention based contextual integration module, which is deeply embedded into the architecture.  Different from \cite{gao2017tall,hendricks2017localizing,wu2018multi,ge2019mac} where context information is simply integrated by feature concatenation, we explicitly measure the different ``contributions'' by leveraging the self-attention technique. Noticeably, our proposed context module operates on the layer which already integrates query and video information. It thus enables our network to ``perceive'' the surrounding predictions and collect reliable contextual evidences before making predictions at the current step. This is different from previous context modeling, which only considers visual context but ignores the impact of language integration. Although LSTMs are also capable of summarizing contextual information, it suffers from the so-called ``gradient vanishing/exploding'' problem and could fail to memorize information for long segments. The proposed contextual model, however, shortens the path for remote elements and effectively aggregates useful contexts in the video.

To summarize, our main contributions are two-folds. First, we address the problem of temporally grounding language queries in videos with a simple yet effective boundary-aware approach, which effectively improves grounding precision in an end-to-end manner. Second, to better detect semantic boundaries, a self attention based module is designed to collect contextual clues. Based on interaction output of both language and video, it explicitly measures the contributions from different contextual elements. Our proposed contextual boundary-aware model (named as CBP) achieves compelling performance on three public datasets.

\setcounter{secnumdepth}{2} 
\section{Related Work}

The interdisciplinary research topics of vision and language have long been explored. Among them we emphasize the following two most relevant topics to our paper: grounding language queries in images, and grounding language queries in videos.

\subsection{Grounding Language Queries in Images}

Grounding language queries in images, also known as ``grounding referring expressions in images'', is to spatially localize the image region corresponding to a given language query. Most work follows the standard pipeline, which first generates candidate image regions using image proposal method like \cite{ren2015faster}, then finds the matched one to the given query. In \cite{mao2016generation,hu2016natural,rohrbach2016grounding}, the target image regions were extracted based on description reconstruction error or probabilities. Some studies consider incorporating contextual information into the retrieval model \cite{hu2016natural,yu2016modeling,chen2017query,chen2017msrc,zhang2018grounding}. These ``contexts'' include global contexts \cite{hu2016natural}, and contexts from other candidate regions \cite{yu2016modeling,chen2017msrc,chen2017query,zhang2018grounding}. \cite{wang2016structured} explored not only region-phrase relationship, but also modeled region-region and phrase-phrase structures. Some other methods exploit attention modeling in queries, images, or object proposals \cite{endo2017attention,yu2018mattnet,deng2018visual}.

\subsection{Grounding Language Queries in Videos}

Temporally video grounding aims at extracting the corresponding video segment to a given language query. Early studies focus on constrained scenarios such as autonomous driving \cite{lin2014visual}, or constrained setting such as alignment of multiple sentences \cite{bojanowski2015weakly}. Recently, \cite{gao2017tall} and \cite{hendricks2017localizing} extended the task to more general scenarios. \cite{gao2017tall} proposed to jointly model video clips and text queries using multi-modal operations, then alignment scores and location offsets were predicted based on the multi-model representation. \cite{hendricks2017localizing} proposed to embed both modalities into a common space and minimize the squared distances. Both \cite{gao2017tall} and \cite{hendricks2017localizing} exploited temporal visual contexts for localization. \cite{wu2018multi} integrated multiple interactions between different modalities and proposed Multi-modal Circulant Fusion. \cite{liu2018attentive} designed a memory attention network to enhance the visual features.
To avoid redundant computation caused by sliding windows, \cite{chen2018temporally} dynamically matches language and video, and generates grounding results in one single pass. 
\cite{liu2018temporal} designed a temporal modular network that can exploit underlying language structure. \cite{ge2019mac} proposed to mine semantic activity concepts to enhance the temporal grounding task. \cite{xu2019multilevel} followed a two-stage pipeline to retrieve video clips. They first generated query-specific proposals from the videos, then leveraged caption reconstruction for training. In \cite{chen2019semantic}, a visual concept based approach was proposed to generate proposals, followed by proposal evaluation and refinement. \cite{wang2019language,hahn2019tripping} explored reinforcement learning to find the corresponding segments to language queries.

\begin{figure}[t]
\centering
\includegraphics [width=1.0\linewidth]{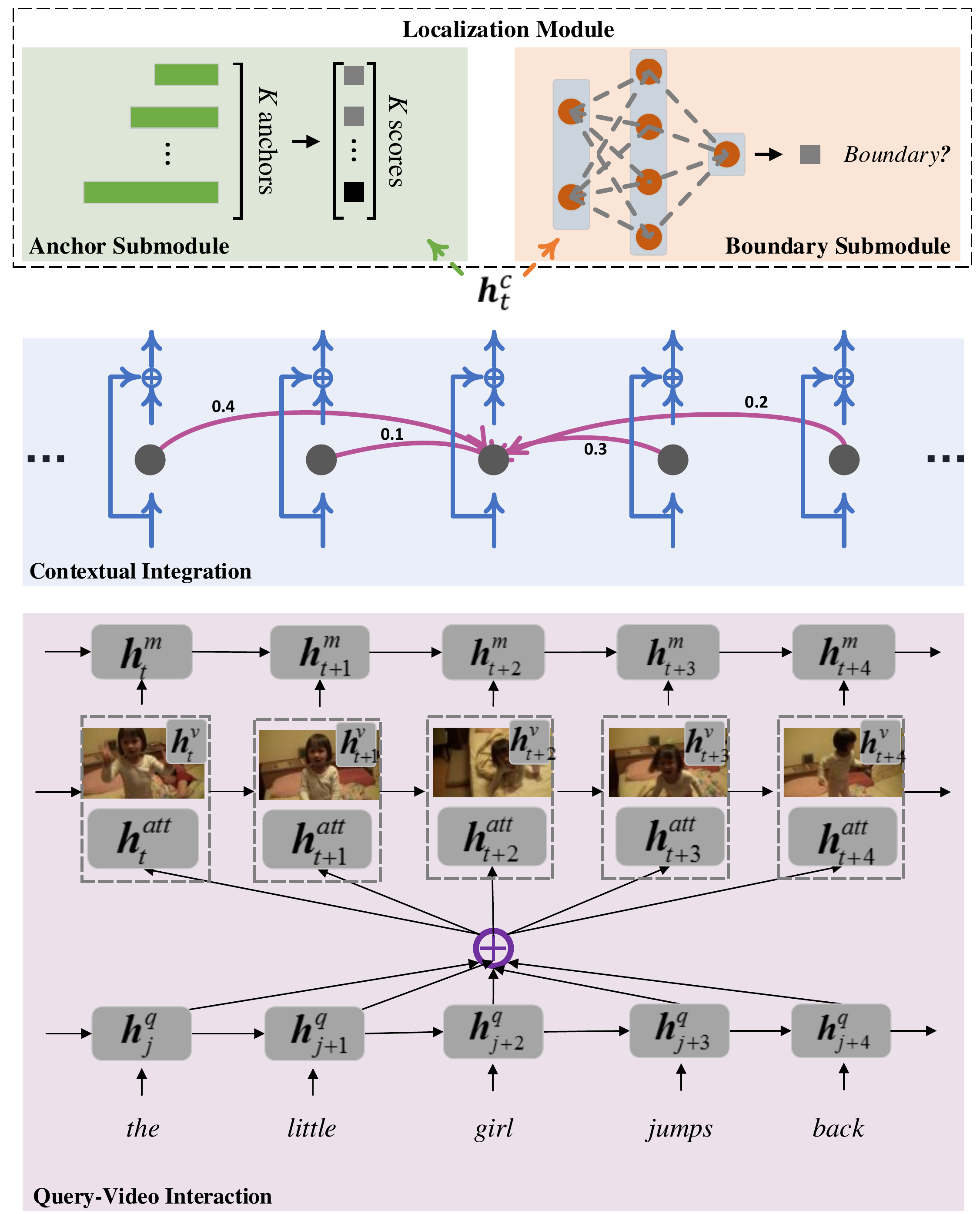}
\caption{The main framework of our proposed method: Contextual Boundary-aware Prediction (CBP). It composes of three modules: a query-video interaction module to deeply integrate language query and video information, a contextual integration module to collect localization clues from neighboring elements, and a localization module to output segments. The localization module consists of an anchor submodule and a boundary submodule.}
\label{figure_method}
\end{figure}

\setcounter{secnumdepth}{3} 
\section{Proposed Method}
\label{method}

In this section we introduce our main framework for temporally grounding queries in videos, as shown in Figure \ref{figure_method}. Our model consists of three main components: the query-video interaction module, the contextual integration module, and the localization module. The three components are deeply integrated and thus enable end-to-end training.

\subsection{Problem Formulation}

We denote a video as a sequence of frames ${X}=\{\mathit{x}_1, \mathit{x}_2, ..., \mathit{x}_L\}$. Each video is associated with a set of annotations: $\{(\mathit{s}_{j}, \mathit{t}_{j}^{s}, \mathit{t}_{j}^{e})\}$, where 
$\mathit{s}_{j}$, $\mathit{t}_{j}^{s}$, $\mathit{t}_{j}^{e}$ denote the query sentence, the start and end time of the annotated segment, respectively. Given the input video and the sentence query, our task is to localize the target segment. Each video is represented as a sequence of features ${V}=\{\mathit{v}_{t}\}_{t=1}^{T}$. The sentence query is represented by ${Q}=\{\mathit{q}_{j}\}_{j=1}^{N}$.

\subsection{Query-Video Interaction Module}

Intrinsically both videos and sentence queries are sequential signals. We incorporate Match-LSTM \cite{wang2016learning,chen2018temporally} as our backbone network to learn vision-language interaction. The Match-LSTM composes of three LSTM \cite{hochreiter1997long} layers. The first LSTM incorporates textual information (denoted as ``query LSTM''). The second LSTM encodes video motion and long-term dependencies from the input video (denoted as ``video LSTM''). The third LSTM is responsible for summarizing video and language elements (denoted as ``interaction LSTM''). The output states of the three LSTMs are $H^q = \{\mathbf{h}_j^q\}$, $H^v = \{\mathbf{h}_t^v\}$, and $H^m = \{\mathbf{h}_t^m\}$, respectively.

As shown in Figure \ref{figure_method}, each video frame is attentively matched to different words from a query:
\begin{align}
\label{equation_word_attetion}
r_{tj} &= \mathbf{w}_r^T \cdot \tanh (W_s \mathbf{h}_{j}^{q} + W_v \mathbf{h}_{t}^{v} + W_m \mathbf{h}_{t}^{m} + \mathbf{b}_r),\\
\alpha_{tj} &= \exp(r_{tj})/\sum_{k=1}^N \exp(r_{tk}), \\
\mathbf{h}_{t}^{att} &= \sum_{j=1}^{N} (\alpha_{tj} \cdot \mathbf{h}_j^{q}), \\
\mathbf{h}_{t+1}^{m} &= \mathit{LSTM}^{m}(\mathbf{h}_{t}^{att} || \mathbf{h}_{t}^{v}, \mathbf{h}_{t}^{m}),
\end{align}
where $\mathbf{h}_{t}^{att}$ is the attended query vector, which relies on current video LSTM state and interaction LSTM state. The attended query vector is concatenated (``$||$'') with the video state ($\mathbf{h}_{t}^{v}$) to serve as input to the interaction LSTM to obtain next state $\mathbf{h}_{t+1}^{m}$.

By the above integration, we deeply summarize and integrate the query and the video.

\subsection{Contextual Integration Module}

To better capture the boundary information corresponding to the starting or ending of an activity, we explore contextual integration by leveraging the self attention technique \cite{vaswani2017attention} on top of the Match-LSTM. Different from pure visual contextual integration \cite{gao2017tall,hendricks2017localizing,hendricks2018localizing,ge2019mac,wu2018multi}, our contextual integration module can strengthen and collect useful grounding clues as it operates on the layer which already integrates query and video information. We also explicitly model the different contributions from different ``contexts'' by assigning them with different attention weights. Formally, the input sequence to the contextual integration module is: $H^m=\{\mathbf{h}_{t}^{m}\}_{t=1,2,...,T}$, where $H^m \in \mathbb{R}^{T \times D}$. Since every pair from $H^m$ needs to be matched, we use scaled dot-product operation to perform self attention as it enjoys high computational efficiency. The relevance matrix for $H^m$ is:
\begin{align}
\label{equation_contextual_relance}
Z &= \frac{1}{\sqrt{d}}(H^m W^{Q})(H^m W^{V})^T,
\end{align}
where the projection matrices $W^Q,W^V \in \mathbb{R}^{D \times d}$ and $Z \in \mathbb{R}^{T \times T}$. \textbf{In practice, we keep $W^Q = W^V$ by sharing projection weights at training.} We find it helps improve the performance. The relevance matrix is then normalized to obtain the context weights $\alpha$:
\begin{align}
\label{equation_contextual_attention}
\alpha_{ij} &= \exp(Z_{ij})/\sum_{t=1}^{T} \exp(Z_{it}).
\end{align}
We summarize contextual elements using the learnt attention to obtain:
\begin{align}
\label{equation_context}
\hat{H}^c &= \alpha H^m.
\end{align}
To avoid corrupting temporal dependency of LSTM, $H^m$ and $H^c$ are integrated by concatenation operation:
\begin{align}
\label{equation_contextual_integration}
H^c = \hat{H}^c || H^m.
\end{align}
$H^c = \{\mathbf{h}_t^c\}_{t=1,2,...,T}$ is expected to strengthen reliable contextual evidence for localization. The operation faithfully preserves the temporal dependency of LSTM, which benefits the following prediction procedure.

\subsection{Localization Module}

The traditional anchor prediction focus more on coarse localization by recognizing segment content. We further propose to strengthen fine-grained semantic boundary information with an additional boundary module. The two modules share the common base network and could benefit each other at the training stage.

\vspace{3pt}
\noindent \textbf{Anchor Submodule}. We adopts similar idea as Buch \textit{et al.} \cite{buch2017sst}. We design $\mathit{K}$ anchors to match different temporal durations. Each $\mathbf{h}_{t}^{c}$ in aggregates historical video information from position 0 to position $\mathit{t}$, after query-video integration. Each hidden state $\mathbf{h}_{t}^{c}$ will be fed into ${K}$ independent binary classifiers and produces ${K}$ confidence scores $\mathit{C_t=\big\{{c_t}^{i}\big\}}_{i=1,...,{K}}$ indicating the probabilities of $K$ segments specified by $\mathit{S_t=\big\{{s_t}^{i}\big\}}_{i=1,...,{K}}$. $\mathit{{s_t}^{i}}$ denotes a video clip with end time as $t$ and start time as $t-\mathit{l}_i$, where $\{\mathit{l}_i\}_{i=1}^K$ is the lengths of the predefined $K$ anchors. The segment scores ${C_t}$ are calculated by:
\begin{align}
\label{equation_score_sigmoid}
{C_t} = \sigma({W_c} \mathbf{h}_{t}^{c} + \mathbf{b}_c),
\end{align}
where $\sigma$ denotes \textit{sigmoid} nonlinearity. ${W_c}$, $\mathbf{b}_c$ are shared across all time steps.

\vspace{3pt}
\noindent \textbf{Boundary Submodule}. Except for the anchor prediction, we also design a parallel branch to predict boundaries of segments. The idea of boundary modeling is simple. We take $\mathbf{h}_{t}^{c}$ as an indication of whether there is a semantic boundary at position $\mathit{t}$. Specifically, a binary classifier is trained with $\mathbf{h}_{t}^{c}$ as input. The output boundary score for current position $\mathit{t}$ is:
\begin{align}
\label{equation_boundary_classifier}
{B_t} = \sigma({W_b} \mathbf{h}_{t}^{c} + \mathbf{b}_b),
\end{align}
which measures how confident the LSTM is going through a semantic boundary. Intuitively, by comparing with its memory (historical video information), the LSTM decides whether the current step is a semantic boundary corresponding the start/end time of an activity (annotated segment).

\subsection{Training}

There are two main losses corresponding to the above two output modules.

\vspace{3pt}
\noindent \textbf{Anchor Loss}. Following \cite{buch2017sst}, the anchor labels $y_t$ ($K$-dim 0-1 vector) at time step $t$ is determined by overlap threshold $\theta=0.5$. We adopt weighted multi-label cross entropy as anchor loss $\mathcal{L}_{a}$. For a video $X$ at time $t$:
\begin{align}
\label{equation_anchor_loss}
\mathcal{L}_{a}(c,t,X,y) = -\sum_{i=1}^K w_0^i y_t^i log c_t^i + w_1^i (1-y_t^i) log (1-c_t^i),
\end{align}
where $w_0^i$, $w_1^i$ are determined based on the numbers of positive and negative samples.

\vspace{3pt}
\noindent \textbf{Boundary Loss}. Assume the training sample $V=\{v_i\}_{i=1}^T$ is associated with ground truth boundary labels $\{z_t\}_{t=1}^T$. The boundary loss is given by:
\begin{align}
\label{equation_boundary_loss}
\mathcal{L}_{b}(t,X,z) = w_{pos} z_t log b_t + w_{neg} (1-z_t) log (1-b_t),
\end{align}
where $b_t \in B_t$ is the boundary prediction score at temporal position $\mathit{t}$, $w_{pos}$ and $w_{neg}$ are positive/negative weights.

\vspace{3pt}
\noindent \textbf{Joint Training}. We balance the anchor loss and the boundary loss by:
\begin{align}
\label{equation_total_loss}
\mathcal{L} = \mathcal{L}_{a} + \lambda \times \mathcal{L}_{b}.
\end{align}
$\lambda$ is determined by cross validation to balance the two loss terms. The CBP network can be trained in an end-to-end manner by minimizing the total loss $\mathcal{L}$.

\subsection{Boundary-modulated Anchor Prediction}

At inference stage, we calculate $\mathit{K}$ anchor scores $\mathit{C}_t \in \mathbb{C}$ and boundary scores $\mathit{B}_t \in \mathbb{B}$ for each video temporal location $\mathit{t} \in \{1, 2, \dots, \mathit{T}\}$. 

\begin{figure}[t]
\begin{center}
\includegraphics [width=0.85\linewidth]{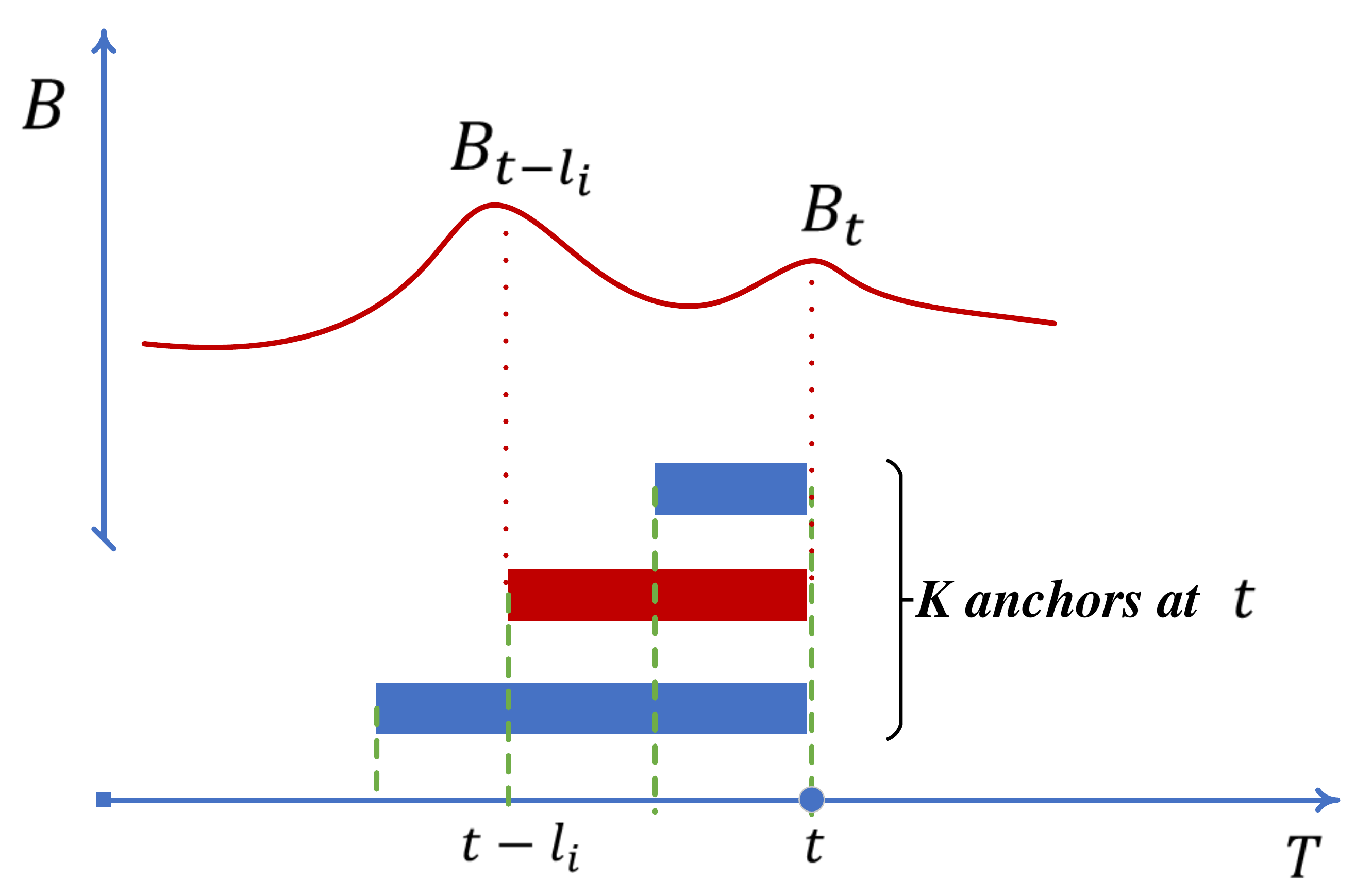}
\vspace{-5pt}
\caption{Local Boundary Score Fusion. The boundary prediction (red curve) helps modulate the score of each anchor. In this case the anchor in red will be selected as the most matched one.}
\label{figure_inference}
\end{center}
\end{figure}

\vspace{3pt}
\noindent \textbf{Local Boundary Score Fusion}. As illustrated in Section \ref{introduction}, the anchor module cannot well reflect boundary information and can produce high scores for many segments that have overlap with the ground truth segment. To precisely localize the target segment, we first apply local score fusion to combine both anchor scores and boundary scores at temporal location $\mathit{t}$. The new scores for the $\mathit{i}$-th anchor at time step $\mathit{t}$ is:
\begin{align}
\label{equation_new_score}
\hat{\mathit{c}}_t^i = \mathit{c}_t^i + 0.5 \times (\mathit{B}_{t-l_i} + \mathit{B}_t),
\end{align}
where $\mathit{c}_t^i \in \mathit{C}_t = \big\{{c_t}^{i}\big\}_{i=1,...,{K}}$. By Equation \eqref{equation_new_score}, we obtain new scores $\hat{\mathit{C}}_t = \big\{\hat{\mathit{c}}_t^i\big\}_{i=1,...,K}$ at each time step $\mathit{t}$. As illustrated in Figure \ref{figure_inference}, we adjust the score of each anchor by taking its start boundary and end boundary into consideration.

\vspace{3pt}
\noindent \textbf{Global Score Ranking}. The final segment scores for a video are $\hat{\mathbb{C}} = \big\{\hat{\mathit{C}}_t\big\}_{t = 1, 2, \dots, \mathit{T}}$. $\mathit{M}$ candidate segments with highest scores are selected and NMS (Non-Maximum Suppression) is performed to further remove redundant candidates. Please note that NMS does not affect top-1 result.

\begin{table*}
{\footnotesize
\caption{Performance comparison on TACoS dataset. All results are reported in percentage (\%).}
\label{table_tacos_performance}
\begin{center}
\begin{tabular} {cccccccc}
\hline
Method & R@1 & R@1 & R@1 & R@5 & R@5 & R@5 & mIoU \\
& IoU=0.7 & IoU=0.5 & IoU=0.3 & IoU=0.7 & IoU=0.5 & IoU=0.3 & \\
\hline\hline
Random Anchor & 0.22 & 0.56 & 2.23 & 1.10 & 3.55 & 9.77 & 1.89 \\
\hline
VSA-RNN \cite{karpathy2015deep} & - & 4.78 & 6.91 & - & 9.10 & 13.90 & - \\
\hline
VSA-STV \cite{karpathy2015deep} & - & 7.56 & 10.77 & - & 15.50 & 23.92 & - \\
\hline
CTRL \cite{gao2017tall} & 6.96 & 13.30 & 18.32 & 15.33 & 25.42 & 36.69 & 11.98 \\
\hline
MCF \cite{wu2018multi} & - & 12.53 & 18.64 & - & 24.73 & 37.13 & - \\
\hline
ACRN \cite{liu2018attentive} & - & 14.62 & 19.52 & - & 24.88 & 34.97 & -\\
\hline
TGN \cite{chen2018temporally}  & 11.88 & 18.90 & 21.77 & 15.26 & 31.02 & 39.06 & 17.93\\
\hline
SM-RL \cite{wang2019language} & - & 15.95 & 20.25 & - & 27.84 & 38.47 & -\\
\hline
TripNet \cite{hahn2019tripping} & 9.52 & 19.17 & 23.95 & - & - & - & -\\
\hline
SAP \cite{chen2019semantic} & - & 18.24 & - & - & 28.11 & - & -\\
\hline
ACL \cite{ge2019mac} & - & 20.01 & 24.17 & - & 30.66 & 42.15 & -\\
\hline
CBP (ours) & \textbf{19.10} & \textbf{24.79} & \textbf{27.31} & \textbf{25.59} & \textbf{37.40} & \textbf{43.64} & \textbf{21.59}\\
\hline
\end{tabular}
\end{center}
}
\vspace{-15pt}
\end{table*}

\begin{table*}
{\footnotesize
\caption{Ablation study on TACoS dataset. All results are reported in percentage (\%).}
\label{table_ablation_performance}
\begin{center}
\begin{tabular} {cccccccc}
\hline
Method & R@1 & R@1 & R@1 & R@5 & R@5 & R@5 & mIoU\\
& IoU=0.7 & IoU=0.5 & IoU=0.3 & IoU=0.7 & IoU=0.5 & IoU=0.3 &\\
\hline\hline
CBP baseline \cite{chen2018temporally} & 11.88 & 20.21 & 25.13 & 15.26 & 30.86 & 38.80 & 17.93\\
\hline
+ Boundary & 16.02 & 22.26 & 25.52 & 22.90 & 34.90 & 41.76 & 19.46\\
\hline
+ Boundary, + Context (full model) & \textbf{19.10} & \textbf{24.79} & \textbf{27.31} & \textbf{25.59} & \textbf{37.40} & \textbf{43.64} & \textbf{21.59}\\
\hline\hline
Replace: Concat-Context & 18.37 & 22.97 & 24.88 & 25.77 & 36.42 & 43.35 & 19.98\\
\hline
Replace: Global-Context & 16.56 & 21.21 & 25.01 & 23.19 & 35.17 & 43.18 & 19.87\\
\hline
Replace: Offset-Reg & 17.68 & 24.69 & 27.31 & 22.73 & 36.08 & 42.59 & 20.79\\
\hline
\end{tabular}
\end{center}
}
\vspace{-10pt}
\end{table*}

\setcounter{secnumdepth}{4} 
\section{Experiments}
\label{experiments}

We conduct extensive experiments on three public datasets: TACoS \cite{regneri2013grounding}, Charades-STA \cite{gao2017tall}, and ActivityNet Captions \cite{krishna2017dense}. For fair comparison, we use the same settings for all baselines, including initial learning rate, segment sampling, NMS threshold, and other hyper-parameters.

\subsection{Datasets}

\noindent \textbf{TACoS}. TACoS is widely used on this task. The videos from TACoS were collected from cooking scenarios. They are around 7 minutes on average. The same split as \cite{gao2017tall} is used, which includes 10146, 4589, 4083 query-segment pairs for training, validation and testing.

\vspace{3pt}
\noindent \textbf{Charades-STA}. Charades-STA was built on Charades dataset \cite{sigurdsson2016hollywood}, which focus on indoor activities. The temporal annotations of Charades-STA were generated in a semi-automatic way, which involved sentence decomposition, keyword matching, and human check. The videos are 30 seconds on average. The train/test split is 12408/3720.

\vspace{3pt}
\noindent \textbf{ActivityNet Captions}. ActivityNet Captions was built on ActivityNet v1.3 dataset \cite{caba2015activitynet}. The videos are 2 minutes on average. Different from the above three datasets, the annotated video clips in this dataset have much larger variation, ranging from several seconds to over 3 minutes. Since the test split is withheld for competition, we merge the two validation subsets ``val\_1'', ``val\_2'' as our test split, as \cite{chen2018temporally}. The numbers of query-segment pairs for train/test split are thus 37421 and 34536.

\subsection{Metrics}

Following prior work, we mainly adopt ``R@$\mathit{N}$, IoU=$\mathit{\theta}$'' and ``mIoU'' as the evaluation metrics. ``R@$\mathit{N}$, IoU=$\mathit{\theta}$'' represents the percentage of top $\mathit{N}$ results that have at least one segment with higher IoU (Intersection over Union) than $\mathit{\theta}$. ``mIoU'' computes the average IoU of top 1 result with ground truth segment over all testing queries.

\subsection{Implementation Details}

For fair comparison, C3D \cite{tran2015learning} features are adopted for all compared methods. Each word from the query is represented by GloVe \cite{pennington2014glove} word embedding vectors pre-trained on Common Crawl. We set hidden neuron size of LSTM to 512.

We generally design the $K$ anchors to cover at least 95\% of training segments. Therefore, we empirically set $K$ to 32, 20 and 100 for TACoS, Charades-STA and ActivityNet Captions, respectively. The NMS thresholds are 0.3, 0.55 and 0.55, respectively.

\begin{table*}
{\footnotesize
\caption{Performance comparison on ActivityNet Captions. All results are reported in percentage (\%).}
\label{table_activitynet_captions_performance}
\begin{center}
\begin{tabular} {cccccccc}
\hline
Method & R@1 & R@1 & R@1 & R@5 & R@5 & R@5 & mIoU\\
& IoU=0.7 & IoU=0.5 & IoU=0.3 & IoU=0.7 & IoU=0.5 & IoU=0.3 &\\
\hline\hline
Random Anchor & 4.54 & 13.28 & 26.64 & 17.95 & 43.40 & 63.65 & 18.40\\
\hline
TGN \cite{chen2018temporally}  & 11.86 & 27.93 & 43.81 & 24.84 & 44.20 & 54.56 & 29.17\\
\hline
Xu \textit{et al.} \cite{xu2019multilevel} & 13.60 & 27.70 & 45.30 & 38.30 & 59.20 & 75.70 & - \\
\hline
TripNet \cite{hahn2019tripping} & 13.93 & 32.19 & 48.42 & - & - & - & -\\
\hline
CBP (ours) & \textbf{17.80} & \textbf{35.76} & \textbf{54.30} & \textbf{46.20} & \textbf{65.89} & \textbf{77.63} & \textbf{36.85}\\
\hline
\end{tabular}
\end{center}
}
\vspace{-10pt}
\end{table*}

\subsection{Compared Methods}

We compare our proposed CBP against the following methods: \textbf{Random Anchor}: the confidence score for each anchor is randomly generated, followed by NMS. \textbf{VSA-RNN} \cite{karpathy2015deep}: visual-semantic alignment with LSTM. \textbf{VSA-STV} \cite{karpathy2015deep}: similar as VSA-RNN, except using skip-thought vectors \cite{kiros2015skip} as query representations. \textbf{CTRL} \cite{gao2017tall}: Cross-model Temporal Regression Localizer. \textbf{ACRN} \cite{liu2018attentive}: Attentive Cross-Model Retrieval Network. \textbf{TGN} \cite{chen2018temporally}: Temporal GroundNet. \textbf{MCF} \cite{wu2018multi}: Multi-modal Circulant Fusion. \textbf{ACL} \cite{ge2019mac}: Activity Concepts based Localizer. \textbf{Xu \textit{et al.}} \cite{xu2019multilevel}: a two-stage method (generation + reranking) exploiting re-captioning. \textbf{SAP} \cite{chen2019semantic}: a two-stage approach based on visual concept grouping. \textbf{SM-RL} \cite{wang2019language}: based on reinforcement learning. \textbf{TripNet} \cite{hahn2019tripping}: leverages RL to perform efficient grounding.

\subsection{Comparison with State-of-the-Arts}
\label{comparison_all}

\noindent \textbf{TACoS.} Table \ref{table_tacos_performance} summarizes performances of different approaches on the test split of TACoS. ``Random Anchor'' is a stronger baseline than uniform random as it eliminates candidates with ``impossible'' durations. However, it achieves very low recalls on all the metrics, indicating that it is quite challenging to accurately localize the desired segment on TACoS. As shown in Table \ref{table_tacos_performance}, the performance generally degenerates for all the methods when IoU gets higher. VSA-RNN and VSA-STV can only achieve unsatisfactory performance compared to the others, mainly because they do not exploit any contextual information for localization. CTRL \cite{gao2017tall}, MCF \cite{wu2018multi}, ACRN \cite{liu2018attentive}, TripNet \cite{hahn2019tripping} and ACL \cite{ge2019mac} use sliding windows to match sentences and video segments, while TGN \cite{chen2018temporally}, SM-RL \cite{wang2019language} and our proposed method CBP adopt LSTMs to eliminate the need of sliding windows. Most sliding window based approaches perform inferior to the single-stream methods (TGN, SM-RL, CBP). ACL \cite{ge2019mac} and SAP \cite{chen2019semantic} perform better than other sliding-window based methods, thanks to the detected visual concepts. Finally, the proposed CBP outperforms all other methods on all the metrics. Noticeably, CBP maintains much better recall rates at high IoUs. For example, for the important metric ``R@1, IoU=0.7'' which indicates high precision, CBP outperforms the others with over 60\% relative gain. This is because CBP is able to generate boundary-aware predictions to match the ground-truth segments more precisely.

\vspace{3pt}
\noindent \textbf{Charades-STA.} The results on Charades-STA are shown in Table \ref{table_charades_performance}. Compared to TACoS dataset, the annotated segments from Charades-STA have a much larger coverage ratio in the video. Therefore, ``Random Anchor'' has much higher recall rates (e.g., 14.65 \textit{vs} 0.22 for ``R@1, IoU=0.5''). We notice that for ``R@5, IoU=0.5'', ``Random Anchor'' obtains a surprisingly high recall (54.35\%). Therefore, we argue that it is better to compare different methods at high IoUs (IoU=0.7 or even higher) on this dataset. Xu \textit{et al.} \cite{xu2019multilevel} leverages multiple useful techniques to enhance the grounding performance, and its results are better than CTRL \cite{gao2017tall}, ACL \cite{ge2019mac}, SAP \cite{chen2019semantic}, SM-RL \cite{wang2019language} and TripNet \cite{hahn2019tripping}. For the important metric ``R@1, IoU=0.7'', our method obtains a recall of 18.87\%, surpassing the previous best result (15.80\%). For the metric ``R@5, IoU=0.5'', Xu \textit{et al.} achieves better recall. One possible reason is that our model finds more false positive boundaries on this dataset.

\begin{table}
{\scriptsize
\caption{Performance comparison on Charades-STA dataset. All results are reported in percentage (\%).}
\label{table_charades_performance}
\begin{center}
\begin{tabular} {cccccc}
\hline
Method & R@1 & R@1 & R@5 & R@5 & mIoU\\
& IoU=0.7 & IoU=0.5 & IoU=0.7 & IoU=0.5 &\\
\hline\hline
Random Anchor & 3.95 & 14.65 & 20.65 & 54.35 & 20.38 \\
\hline
VSA-RNN & 4.32 & 10.50 & 20.21 & 48.43 & - \\
\hline
VSA-STV & 5.81 & 16.91 & 23.58 & 53.89 & - \\
\hline
CTRL & 7.15 & 21.42 & 26.91 & 59.11 & -\\
\hline
ACL & 12.20 & 30.48 & 35.13 & 64.84 & 33.84\\
\hline
SAP & 13.36 & 27.42 & 38.15 & 66.37 & -\\
\hline
SM-RL & 11.17 & 24.36 & 32.08 & 61.25 & -\\
\hline
TripNet & 14.50 & 36.61 & - & - & -\\
\hline
Xu \textit{et al.} & 15.80 & 35.60 & 45.40 & \textbf{79.40} & - \\
\hline
CBP (ours) & \textbf{18.87} & \textbf{36.80} & \textbf{50.19} & 70.94 & \textbf{35.74}\\
\hline
\end{tabular}
\end{center}
}
\vspace{-15pt}
\end{table}

\vspace{3pt}
\noindent \textbf{ActivityNet Captions.} As can been seen from Table \ref{table_activitynet_captions_performance}, our CBP surpasses both TGN \cite{chen2018temporally} and Xu \textit{et al.} \cite{xu2019multilevel} on all the metrics with a clear margin. The proposed CBP obtains 17.04\% at ``R@1, IoU=0.7'' while Xu \textit{et al.} and TripNet can only achieves 13.60\% and 13.93\% respectively. This provides strong evidences on the superiority of the proposed CBP. Similar to Charades-STA, many of the annotated segments on ActivityNet Captions dataset are long compared to the video duration. Therefore, for low IoUs (e.g., IoU=0.3), many approaches perform similarly to the ``Random Anchor'' baseline. We also notice that CBP achieves less relative improvement over Xu \textit{et al.} and TripNet for lower IoUs (e.g., IoU=0.3). This is because our model focus more on localization precision.

\begin{figure*}
\centering
\includegraphics [width=0.85\linewidth]{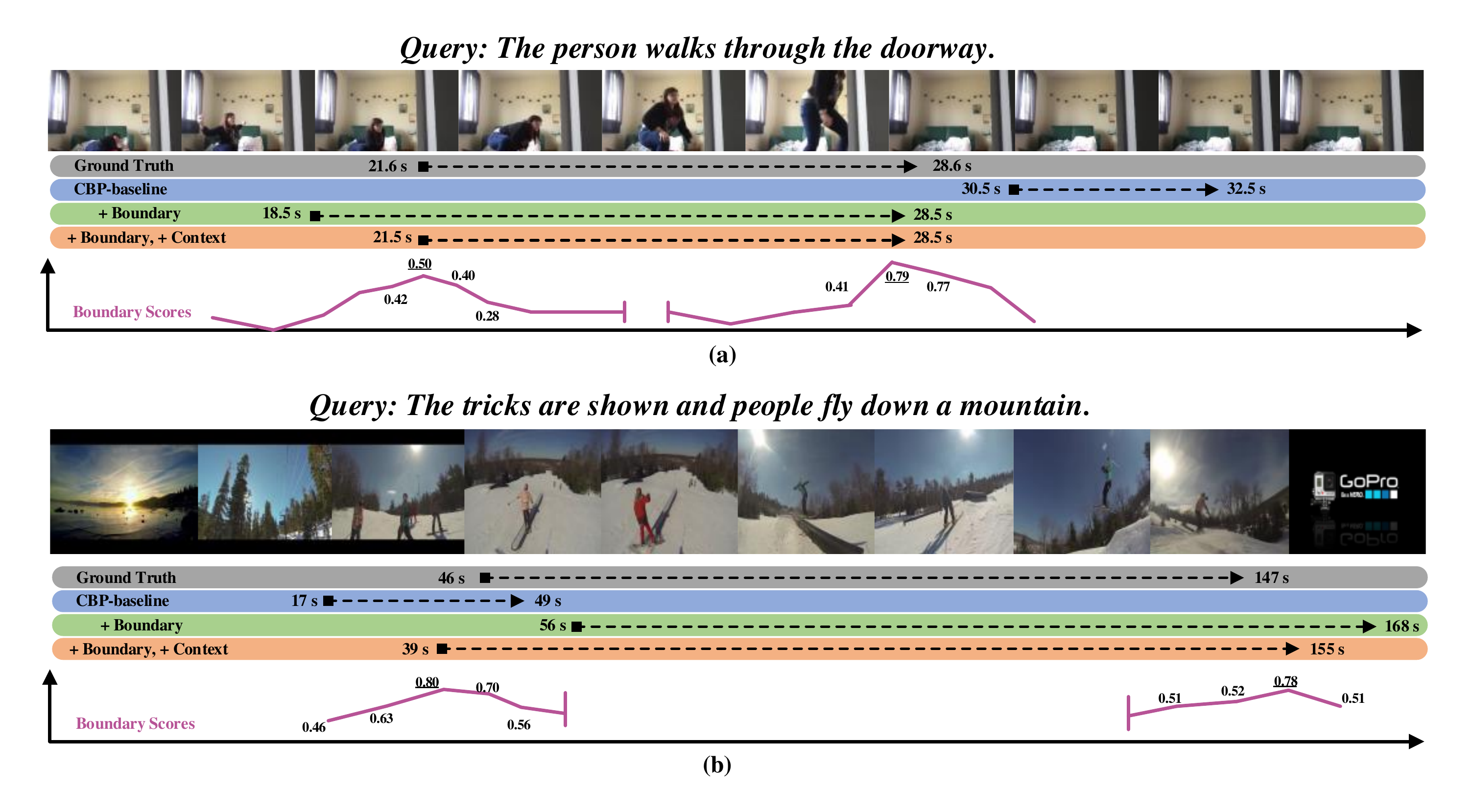}
\vspace{-8pt}
\caption{Prediction examples of our CBP and the baselines. The boundary scores are computed using our full CBP model.}
\label{figure_cases}
\end{figure*}

\subsection{Ablation Study}

To evaluate each component of the proposed CBP model, we conduct ablation study on TACoS dataset. The results are shown in Table \ref{table_ablation_performance}. We observe substantial performance improvement when applying the proposed boundary module, especially for the metrics of high IoUs (e.g., ``R@1,IoU=0.7'', ``R@5,IoU=0.7''). This indicates that equipping with the boundary module greatly improve the grounding precision. CBP outperforms all other methods when further integrating the context module (``+ Boundary, + Context''). Moreover, each module of CBP is compared to existing techniques by replacement in order to further verify the effectiveness of the proposal. The first experiment is to replace our proposed self-attention based contextual integration module with the commonly-adopted concatenation-based contextual module \cite{gao2017tall,hendricks2017localizing,wu2018multi,ge2019mac} or the global contextual module \cite{wang2019language,hendricks2017localizing}. The second one is to replace our boundary module with an offset regression branch \cite{gao2017tall,xu2019multilevel,ge2019mac,liu2018attentive}. The performance degeneration observed in Table \ref{table_ablation_performance} verifies the superiority of our proposed modules over their corresponding competitors.

\begin{figure}
\centering
\includegraphics [width=1.00\linewidth]{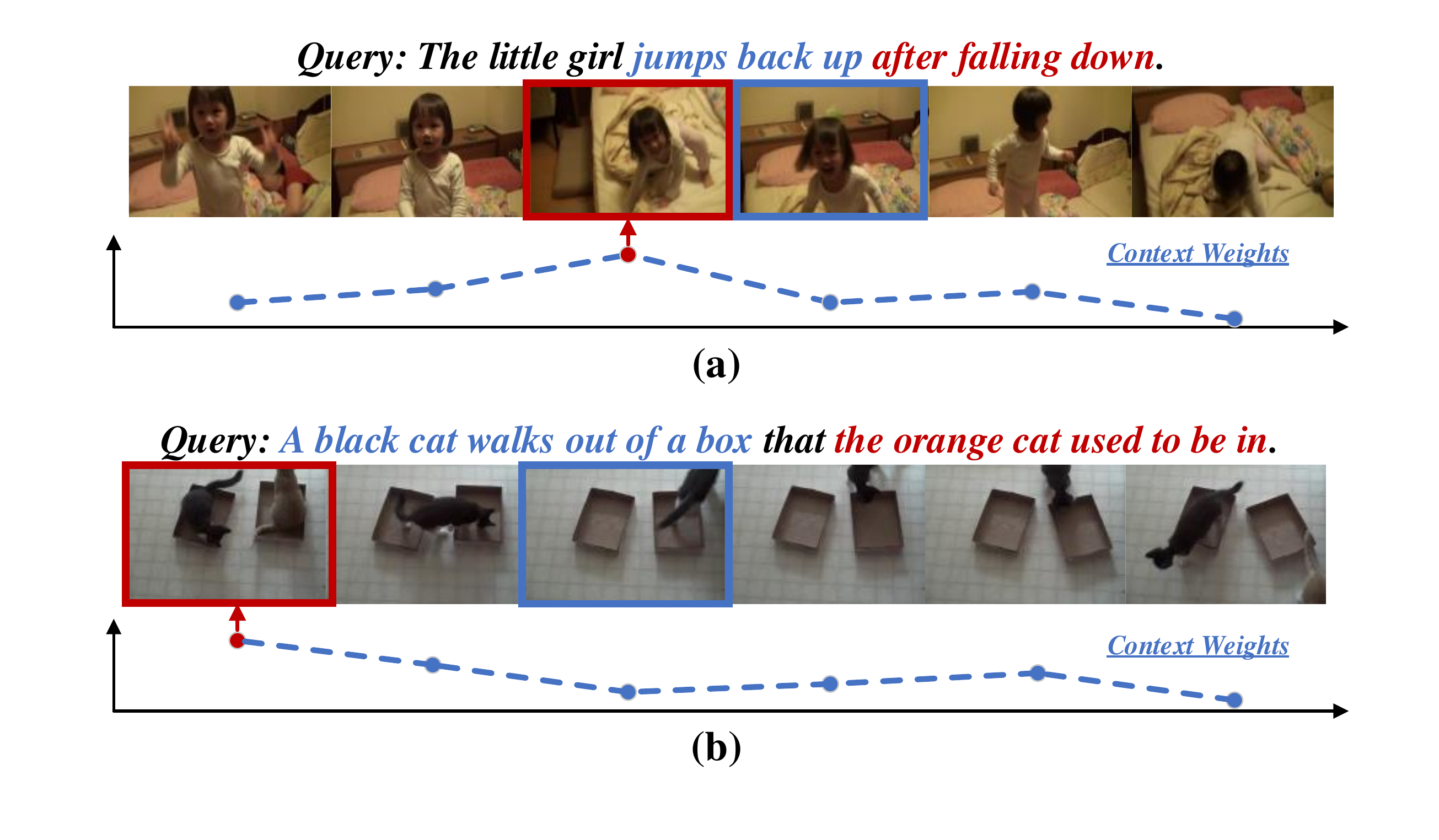}
\vspace{-12pt}
\caption{Visualization of the learnt context weights. Ground-truth segments are outlined in blue boxes. Contextual segments corresponding to the highest context weights are outlined in red boxes.}
\label{figure_context_weights}
\end{figure}

\subsection{Qualitative Analysis}

We provide some qualitative examples to validate the effectiveness of the proposed CBP. As shown in Figure \ref{figure_cases}, the boundary prediction module exploits boundary information and modulates the anchors by combining predictions from both output modules. This makes it perform better than the CBP baseline. By contextual integration, the boundaries of the desired segment can be further recognized. 

We also visualize the learnt context weights in Figure \ref{figure_context_weights}. Each blue box represents the ground-truth segment to be localized and each red box corresponds to the segment with the highest context weight. In Figure \ref{figure_context_weights} (a), our model successfully pinpoint the desired activity ``\texttt{\small{jumps back up}}'' (in blue box) by attending to its precursor action ``\texttt{\small{falling down}}'' (in red box). In Figure \ref{figure_context_weights} (b), to accurately localize the desired segment in blue box, the model resorts to the segment in red box, which shows the visual content of ``\texttt{\small{a box that the orange cat used to be in}}''. We notice that the best context is not necessarily the nearest segment to the queried segment, as evidenced by Figure \ref{figure_context_weights} (b).

\vspace{5pt}
\setcounter{secnumdepth}{5} 
\section{Conclusion}

In this paper, we proposed a contextual boundary-aware model (CBP) to address the task of temporally grounding language queries in videos. Different from most prior work, CBP was built with a single-stream architecture, which processes a video in one single pass. The idea of boundary prediction is simple yet effective. The promising experimental results obtained on three widely-used datasets demonstrated the effectiveness of our model.


\end{document}